\documentclass[letterpaper, 10 pt, conference]{ieeeconf}  

\IEEEoverridecommandlockouts                              

\usepackage{graphics} 
\usepackage{epsfig} 
\usepackage{times} 
\usepackage{amsmath} 

\usepackage[ruled,vlined,linesnumbered]{algorithm2e}
\usepackage{xargs}    
\usepackage{textcomp}
\usepackage{multirow}
\usepackage{lipsum}
\usepackage{subcaption}
\usepackage{xcolor}
\usepackage{bm}
\usepackage{cite}
\usepackage{float}
\usepackage{graphicx}
\usepackage[]{todonotes}    
\usepackage{framed, color} 
\usepackage[normalem]{ulem}

\newcommandx{\aleksi}[2][1=]{\todo[backgroundcolor=green!25,inline,#1]{ALE: #2}\noindent}
\newcommandx{\ville}[2][1=]{\todo[backgroundcolor=blue!25,inline,#1]{V: #2}\noindent}
\newcommandx{\karol}[2][1=]{\todo[backgroundcolor=red!25,inline,#1]{K: #2}\noindent}
\newcommandx{\ali}[2][1=]{\todo[backgroundcolor=yellow!25,inline,#1]{ALI: #2}\noindent}
\SetKwComment{Comment}{$\triangleright$\ }{}

\title{\LARGE \bf
Affordance Learning for End-to-End Visuomotor Robot Control
}

\author{Aleksi H{\"a}m{\"a}l{\"a}inen$^{1}$, Karol Arndt$^{1}$, Ali Ghadirzadeh$^{1,2}$ and Ville Kyrki$^{1}$ 
\thanks{*This work was supported by Academy of Finland grant 313966.}
\thanks{$^{1}$Aalto University, Espoo, Finland
        {\tt\small first.last@aalto.fi}}%
\thanks{$^{2}$KTH Royal Institute of Technology, Stockholm, Sweden}%
}

\begin{document}
\maketitle
\thispagestyle{empty}
\pagestyle{empty}

\begin{abstract}
Training end-to-end deep robot policies requires a lot of domain-, task-, and hardware-specific data, which is often costly to provide.
In this work, we propose to tackle this issue by employing a deep neural network with a modular architecture, consisting of separate perception, policy, and trajectory parts.
Each part of the system is trained fully on synthetic data or in simulation.
The data is exchanged between parts of the system as low-dimensional latent representations of affordances and trajectories.
The performance is then evaluated in a zero-shot transfer scenario using Franka Panda robot arm.
Results demonstrate that a low-dimensional representation of scene affordances extracted from an RGB image is sufficient to successfully train manipulator policies. We also introduce a method for affordance dataset generation, which is easily generalizable to new tasks, objects and environments, and requires no manual pixel labeling.

\end{abstract}

\vspace{-1pt}
\section{Introduction}
\vspace{-1pt}
Recent years have seen widespread research and adoption of deep learning in robotics. 
Convolutional neural networks became the \textit{de facto} standard way of processing visual input, and generative models are starting to see wider adoption as trajectory generators \cite{ghadirzadeh2017deep}.
Data-driven approaches like deep learning make it possible to learn diverse behaviours with little feature engineering \cite{levine2016end} and can be scaled to complex problems if enough data is available.

Despite extensive use of deep learning, a few issues limit its further adoption. Specifically, (i) end-to-end training requires vast amounts of task-specific training data and months of real world experience \cite{levine2017learning}, which are simply unfeasible to provide;
(ii) using task- and domain-specific training data means that changing the task, such as transferring it to another robot or changing the task objective, requires often  entirely new training data, which is costly to obtain \cite{levine2016end};
(iii) the behaviour of deep models can be difficult to predict and interpret, as the intermediate representations learned by the neural network rarely provide meaningful information to humans --- despite recent successes in interpreting neural networks, it is still difficult to gain understanding of why the model behaves the way it does.

\begin{figure}
    \centering
    \begin{subfigure}{0.48\linewidth}
        \centering
        \includegraphics[width=\linewidth]{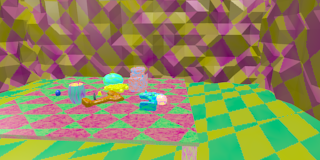}
        \caption{}
        \label{fig:intro_ble}
    \end{subfigure}
    \begin{subfigure}{0.48\linewidth}
        \centering
        \includegraphics[width=\linewidth]{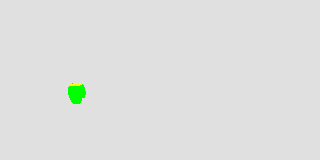}
        \caption{}
        \label{fig:intro_ble_aff}
    \end{subfigure}
    \begin{subfigure}{0.48\linewidth}
        \centering
        \includegraphics[width=\linewidth]{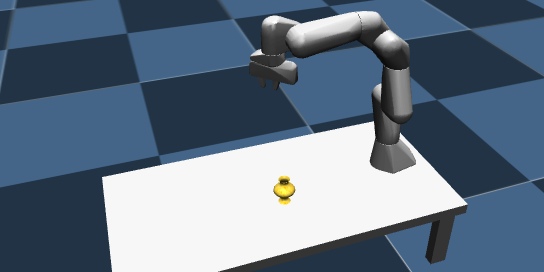}
        \caption{}
        \label{fig:intro_mujoco}
    \end{subfigure}
    \begin{subfigure}{0.48\linewidth}
        \centering
        \includegraphics[width=\linewidth]{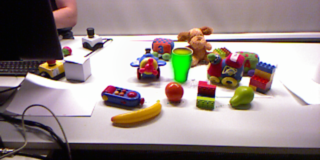}
        \caption{}
        \label{fig:intro_re}
    \end{subfigure}
    \vspace{-3pt}
    \caption{In our experiments, we train an affordance detection model using synthetic scene images (\subref{fig:intro_ble}) with their respective affordance images (\subref{fig:intro_ble_aff}). The intermediate representations of this model are used to train manipulator policies in MuJoCo (\subref{fig:intro_mujoco}). The performance of the system is evaluated in a real world setup (\subref{fig:intro_re}). The cup's affordances, \textit{wrap-grasp} and \textit{contain}, are respectively marked in green and yellow.}
    \label{fig:intro}
\vspace{-14pt}
\end{figure}

In this paper, we propose to address those issues with a modular neural network, which consists of three parts: perception, policy and trajectory generation.
Each part of the system is, on its own, a neural network trained entirely on synthetic data or in simulation, with no real-world adaptation.
Information is exchanged between parts using low-dimensional representations of affordances and trajectories.
By using visual affordances, the intermediate representation captures semantic information of what can be done with each part of each object \cite{gibsontheory}, which provides the system with generality across different tasks. 
\vspace{-0.5pt}

The perception part is a variational encoder-decoder structure \cite{kingma2013autoencoding} trained for affordance detection from synthetic RGB images.
Those images are generated using randomized textures, object shapes, distractor objects and camera viewpoints to maximize the model's capability to generalize to new environments \cite{tobin2017domain, james2017transferring}.
The encoder part outputs a low-dimensional representation of scene affordances.
The trajectory part is a variational autoencoder trained on a set of trajectories.
The policy part, trained in a physics simulator, maps latent affordance representations to latent representations of trajectories, which are then passed to the trajectory decoder.
Using variational autoencoders ensures smoothness of the latent space \cite{Higgins2016EarlyVC}, which, together with the low dimensionality of the latent representation, makes it quick to retrain the policy layers using a small amount of training data.
\vspace{-0.5pt}

We demonstrate that the system achieves good performance in the task of inserting a ball into a container both in a simulated environment and on real hardware (the Franka Panda robotic arm), without the need for any real world  adaptation of the parts (a zero-shot transfer scenario).
We also evaluate its susceptibility to clutter, demonstrating that it can complete the task even in heavily cluttered environments, such as shown in Figure \ref{fig:intro_re}, without having any explicit model of the cup object.
\vspace{-1pt}

We also propose a new method of generating a domain randomized dataset for affordance detection, which allows to automatically generate vast amounts of data for affordance training without the need of hand-labeling pixels.
Perception models trained using this affordance data can then be used both in real world and in simulated environments. The code for generating the dataset and training our models is available online\footnote{
https://github.com/gamleksi/affordancegym}.
\vspace{-1pt}


\vspace{-1pt}
\section{Related work}
\vspace{-1pt}

In this section, we introduce studies related to the problem of visual perception training for control. The three commonly used approaches to address the problem of perception training are: (1) end-to-end training of the perception and control layers in a single neural network architecture, e.g., \cite{levine2016end,devin2018deep,singh2017gplac}, 
(2) perception training using priors and auxiliary tasks, e.g., \cite{finn2016deep, ghadirzadeh2017deep}, and
(3) perception training with labeled affordances, e.g., \cite{song2016learning, nguyen2016detecting, nguyen2017object,do2018affordancenet}. 
\vspace{-0.5pt}

End-to-end training results in task specific robust features when large enough training datasets are provided \cite{levine2016end}. However, even with large datasets, the learned features may not generalize well to similar tasks. 
One approach to solve this issue is to introduce extra loss functions to extract general multitask features. 
A popular solution is to
 \textit{learn low-dimensional state representations} of the visual data \cite{ finn2016deep, ghadirzadeh2017deep,watter2015embed,van2016stable}. 
Encoder-decoder architectures (also known as auto-encoders when the input data is restored at the output) are widely used models for this purpose. The hourglass shape of the network abstracts the input data at the bottleneck (the latent space) based on which the output data can be constructed. 
Spatial autoencoders, introduced by Finn et al., \cite{finn2016deep}, use CNNs to extract a  
low-dimensional feature vector consisting of spatial image points to localize task-related objects in the input image. 
Spatial features are suitable for manipulation tasks as they contain information to move the robot end-effector to desired positions. 
However, the features may not be stable and robust given limited training data \cite{ghadirzadeh2017deep}. 
\vspace{-0.5pt}

In order to improve the robustness and the generalization power of the extracted features, prior knowledge and auxiliary tasks \cite{jonschkowski2015learning,singh2017gplac,devin2018deep,jaderberg2016reinforcement,mirowski2016learning} are introduced as extra loss functions for the training.  
Prior knowledge ranges from physical laws \cite{jonschkowski2015learning} to high-level semantic features \cite{devin2018deep,singh2017gplac}. 
Overview of different priors for the perception training problem can be found in \cite{ghadirzadeh2018sensorimotor} and \cite{lesort2018state}.
We train a low-dimensional representation of visual inputs using variational encoder-decoder structures, which, similar to \cite{watter2015embed, van2016stable},  assigns a normal distribution over the latent space conditioned on the observations. This prevents similar states from being scattered in the latent space, which is a suitable property for control purposes. Furthermore, the features can be generally used for different tasks, since they are trained to extract high-level action affordances of the objects irrespective of the task.
Similar to \cite{ghadirzadeh2017deep}, we also trained a low-dimensional representation of motor trajectories based on variational autoencoders. This further improves the data efficiency of the proposed method. 
\vspace{-0.5pt}

\textit{Learning visual affordances} is an approach to identify a set of functions available to an object given visual observations.
Several recent studies worked on understanding affordances at the image pixel level, i.e., to segment pixels with the same affordances \cite{hassanin2018visual, chuang2018learning, luddecke2017learning, roy2016multi}. 
However, there are few studies that deal with learning affordance representations suitable for control.
\vspace{-0.5pt}

Myers et al. \cite{myers2015affordance} demonstrated that local geometric primitives can be used to detect pixel-level affordances. 
Song et al. \cite{song2016learning} proposed to combine a local affordance extraction method based on the SIFT descriptor, and a global object grasp localization to learn grasp affordances for robotic applications. 
Nguyen et al. \cite{nguyen2016detecting} proposed a CNN encoder-decoder architecture to learn different affordances from RGB-D images. 
In a more recent work, they proposed to use an object detector network to narrow down the image regions at which object affordances are extracted. 
Do et al. \cite{do2018affordancenet} proposed to jointly optimize the object detection and the affordance learning losses in a single feed-forward neural network to speed-up the process in the deployment phase.  
However, these methods do not extract a suitable representation of the action affordances that can be directly used for control purposes. All the methods are based on a post-processing step in which affordance images are further processed to obtain suitable motor actions. 
In our method, we extract pixel-level affordances together with a low-dimensional representation which can be efficiently used for end-to-end visuomotor policy training. 
\vspace{-0.5pt}

A major issue with pixel-wise affordance learning is the intensive image labeling work required to collect training datasets. To alleviate this issue, Srikantha and Gall
\cite{srikantha2016weakly} introduced an expectation-maximization method to learn pixel-level affordances given weakly labeled image data. 
In this work, we propose a different approach to overcome the difficulties of manual pixel-level labeling. We leverage sim-to-real transfer learning based on domain randomization to collect a diverse set of training data which can be used with the real image inputs.  
\vspace{-0.5pt}

\textit{Domain randomization} in simulation is a technique for zero-shot sim-to-real transfer learning by randomizing non-essential aspects of a simulated process. This helps to improve the diversity of the training dataset, which, in turn, helps the extraction of a set of task-relevant features that can be transferred to real world problems.
Our method resembles the work of \cite{tobin2017domain,james2017transferring,james2018sim} in that we render simulated scenes by randomizing textures and appearances of the objects in the scene. An important contribution of our work is that we get large-scale pixel-level affordance labels automatically without extra manual labeling work.

\vspace{-1pt}
\section{Method}
\vspace{-1pt}

\begin{figure}
    \centering
    \includegraphics[width=\linewidth]{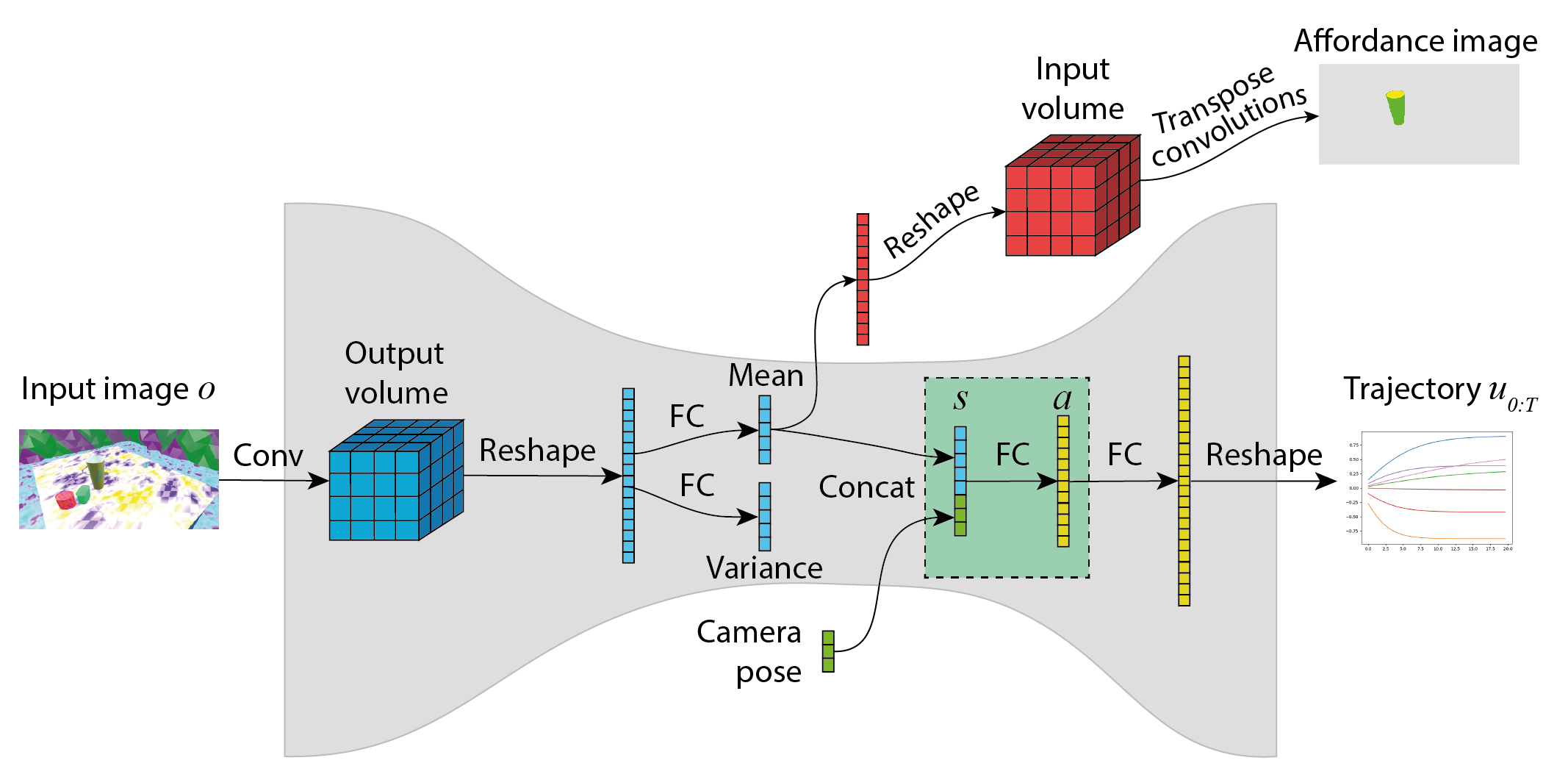}
    \caption{Structural overview of the system. An input image $o$ is first processed by the affordance encoder (blue). The policy part (green rectangle) produces action $a$, which is then decoded to a robot trajectory $u_{0:T}$. The affordance image is also generated by the affordance decoder (red).}
    \label{fig:structure}
    \vspace{-16pt}
\end{figure}

The system consists of three separately trained parts---affordance perception, policy, and trajectory generation, similar to \cite{ghadirzadeh2017deep}.
The perception part encodes an RGB image $o$ to a low-dimensional affordance representation $s$.
The policy part maps the state $s$ to a desired low-dimensional action vector $a$.
Finally, based on the action $a$, the trajectory part generates a trajectory of robot joint positions $u_{0:T}$.
\vspace{-1pt}

This structure is shown in Figure \ref{fig:structure}.
The affordance encoder consists of convolutional and fully connected layers (blue), while the decoder has the inverse structure (fully connected layers followed by transposed convolutions; marked in red).
The yellow blocks represent the policy and trajectory generation parts.
Information about camera pose is added to the latent space before passing it to the policy to allow it to estimate the 3D pose of the target object w.r.t. the camera.
Gray background emphasizes the hourglass shape of the network with a low-dimensional information bottleneck in the middle.
Both perception and trajectory blocks are built using variational encoder-decoder structures introduced in the next section. 
\vspace{-2pt}

\subsection{Variational encoder-decoder structures}
\vspace{-2pt}
To facilitate quick training and to improve data efficiency of the learning method, we use low-dimensional representations of affordances and trajectories as the intermediate steps between each part of the system.
To obtain these low-dimensional representations in an unsupervised way without using task-specific hand-engineered features, we use variational encoders-decoders --- a more general case of a variational autoencoder (VAEs).
\vspace{-1pt}

Autoencoders are hourglass shaped neural networks with a low-dimensional information bottleneck in the middle, which encodes the \textit{latent representation} of the input vector.
Variational autoencoders additionally incorporate a KL-divergence term in their loss functions, which encourages the neighbourhood preservation and disentanglement properties in the latent space \cite{Higgins2016EarlyVC, Burgess2018UnderstandingDI}.
The trade-off between these properties and reconstruction quality can be balanced by the $\beta$ hyperparameter, introduced by \cite{Higgins2016EarlyVC}.
In our method, these properties make it easier for the policy to find a smooth mapping between the affordances and the actions.

In contrast to a typical VAE application, our affordance perception model uses the latent representation to create an affordance map, instead of reconstructing the input.
To emphasize on this, we refer to it as a \textit{variational encoder-decoder structure}.
\vspace{-3pt}

\subsection{Affordance perception}
\vspace{-1pt}

The perception layer extracts affordance information from an observation $o$ and encodes it as a latent space vector $s$.
Some properties, such as textures and the surrounding environment, have no impact on object affordances and can be ignored by the perception model.

To obtain those low-dimensional affordance representations, we use a variational affordance encoder-decoder (VAED) structure, which encodes an RGB image and generates the corresponding pixel-level affordance labeling.
The encoder of VAED consists of a series of convolutional layers followed by one fully connected layer.
The mean and covariance are produced by separate fully connected layers.
The latent vector, drawn from the posterior distribution, is then passed as an input to the decoder.
The decoder has a reversed structure---a fully connected layer followed by a series of transposed convolution layers.
The decoder produces a multi-channel probability map, with each channel corresponding to a specific affordance.
A pixel value in an affordance channel describes how likely that affordance is to occur in that pixel.
The probability output is achieved by using the sigmoid activation function.
We use binary cross entropy as the reconstruction loss.

Training the affordance detection network requires corresponding labeled affordance images.
The target labels are generated in Blender together with the RGB inputs.
We use domain randomization of textures, clutter and lighting to increase the robustness of the model.
Even though artificial textures make the input images look unrealistic, high variation allows the model to generalize to real world environments \cite{tobin2017domain}.
\vspace{-4pt}

\subsection{Trajectory generation}
\vspace{-2pt}

To build a generative model for trajectories, we first train a variational autoencoder on a set of task-specific trajectories, similar to \cite{ghadirzadeh2017deep}.
The decoder block of this autoencoder converts low-dimensional latent trajectory representations of actions $a$ to their corresponding trajectories in joint space $u_{0:T}$.
This model can therefore be used as a task-specific trajectory generator---it converts each latent action $a$ to a trajectory that is useful for accomplishing the given task.
To achieve this, we rely on the observation that trajectories useful for a specific task often exhibit structural similarities, e.g., trajectories useful for pouring liquids into containers consist of motions to a point above the target location followed by a wrist rotation.
We generate such trajectories using MoveIt!, an open source motion planning software.
The encoder and the decoder of the trajectory VAE consist of three fully connected layers each.
We use mean squared error (MSE) as the reconstruction loss.
\vspace{-4pt}

\subsection{Policy layer}
\vspace{-2pt}

A policy part maps the affordance information $s$ to a desired action vector $a$. The policy part includes three fully connected layers.
In addition to the affordance information $s$, camera parameters are included as input to the policy network, which  allows the policy to account for the relative  position of the camera w.r.t the detected affordances.
The policy part can be trained either in a supervised manner or using reinforcement learning.
We utilize supervised learning to train the policy.

\vspace{-2pt}
\section{Experiments}

The goal of the experiments was to study if a low-dimensional affordance representation can be used to successfully train manipulator  policies. In particular, we first evaluated if affordances can be represented in a latent space using the UMD dataset \cite{myers2015affordance}, a standard benchmark for the affordance detection task. 

\begin{figure}
    \centering
    \includegraphics[width=\linewidth]{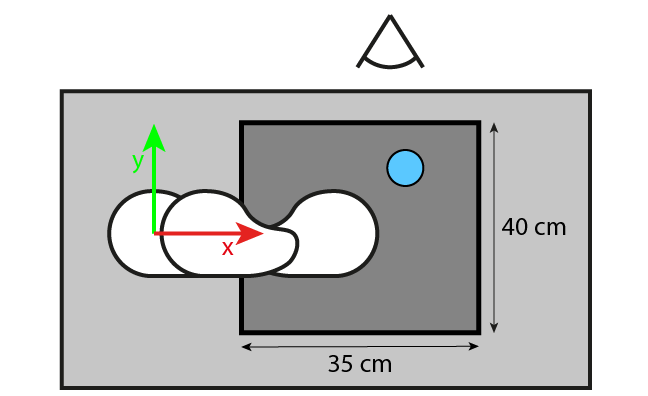}
    \caption{Experimental setup, as viewed from top. The robot is marked in white. The dark gray area shows the workspace and the blue circle an example location of a target container The camera is located on the left side (from the robot's perspective).}
    \label{fig:exp_setup}
\vspace{-3pt}
\end{figure}

To study the approach as a whole, we developed an experimental set-up presented in Figure \ref{fig:exp_setup}.
In this setup, the task of the Franka Panda robot arm was to insert a ball into a container that is located on a table, with the scene observed by an RGB-camera. We used the set-up to evaluate the performance quantified as position accuracy and task success both in simulation and on physical hardware.

\vspace{-3pt}

\subsection{UMD dataset benchmark}
\vspace{-3pt}

\begin{table}
    \caption{Results of the UMD dataset evaluation}
    \label{tab:umd}
    \centering
    \begin{tabular}{c||c|c||c|c}
        \multirow{2}{*}{Affordance} & \multicolumn{2}{c||}{VAED} & \multicolumn{2}{c}{Nguyen et al.} \\
        \cline{2-5}
         & RGB & RGB-D & RGB & RGB-D \\ 
         \hline
         grasp & 0.667 & 0.643 & \textbf{0.719} & 0.714 \\
         cut & 0.720 & 0.654 & \textbf{0.737} & 0.723 \\
         scoop & \textbf{0.760} & 0.758 & 0.744 & 0.757 \\
         contain & \textbf{0.859} & 0.858 & 0.817 & 0.819 \\
         pound & 0.757 & 0.748 & 0.794 & \textbf{0.806} \\
         support & 0.792 & 0.787 & 0.780 & \textbf{0.803} \\
         wrap-grasp & 0.774 & \textbf{0.777} & 0.769 & 0.767 \\
         \hline
         Average & 0.761 & 0.747 & 0.766 & \textbf{0.770} \\
    \end{tabular}
\vspace{-12pt}
\end{table}

We first evaluated the capabilities of variational affordance encoder-decoder to detect object affordances on the UMD dataset~\cite{myers2015affordance}, randomly splitting it into training (70\%) and validation (30\%) data.  
For this evaluation, we used 20 dimensional latent space and KL-divergence penalty $\beta = 4$.
We used $F^w_\beta$ as the evaluation metric~\cite{margolin2014how}.
The results are presented in Table~\ref{tab:umd} and are compared to the CNN-RGB and CNN-RGBD networks introduced in \cite{nguyen2016detecting}.

We can notice that for some affordances --- \textit{contain} and \textit{wrap-grasp} --- our model slightly outperformed the baseline, while for the other affordances the results were the same or slightly worse.
Comparing the RGB and RGB-D results, we can also see that discarding the depth information does not have much impact on the performance.

This evaluation shows that the variational encoder-decoder structure using RGB images can be used for affordance detection.
\vspace{-3pt}

\subsection{Training the perception model}
\vspace{-3pt}

\begin{figure}
 \vspace{5pt}
    \centering
    \includegraphics[width=0.9\linewidth]{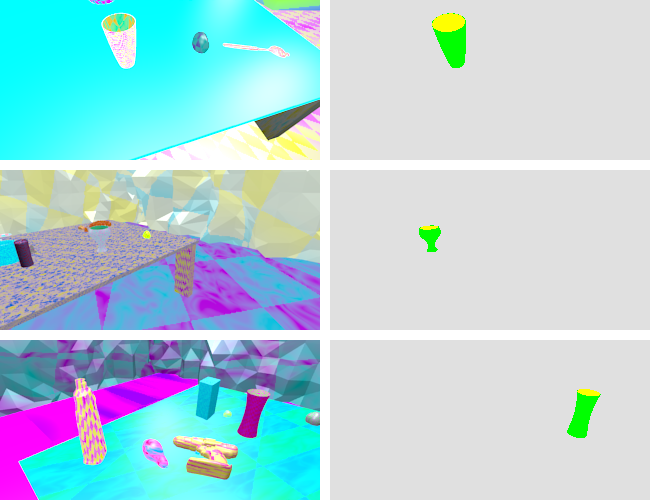}
    \caption{Samples from the affordance dataset. Green indicates the \textit{wrap-grasp} affordance and yellow indicates the \textit{contain} affordance. Gray pixels mean no affordance at all.}
    \label{fig:blender_affo}
  \vspace{-9pt}
\end{figure}

We used Blender's Python API to generate a domain randomized dataset of 1 million images. 
Figure \ref{fig:blender_affo} shows examples from the generated dataset.
The Blender environment was constructed similar to the experimental setup: it included a table with a cup and clutter objects, and the table was surrounded by walls.
Each sample includes a rendered RGB image and a corresponding affordance image, which was used as the target output during training.

A simple cylinder model was built in Blender to represent a cup-like object.
Its shape was randomly generated by smoothly changing its diameter along the height.
Textures of the inner and outer parts of the object were separately randomized.
In our setting, two different affordances occur: the outer part  has a \textit{wrap-grasp} affordance and the inner part has a \textit{contain} affordance.
Clutter objects were located on the table and their affordances were ignored.
In total, 66 clutter objects from the Yale-CMU-Berkeley (YCB) Object and Model set \cite{calli2015benchmarking} were used.

The following features were randomized uniformly within reasonable limits: (1) positions, scales, and textures of the clutter objects and the cup, (2) shape of the cup, (3) texture and scale of the table, (4) camera pose, (5) the number of clutter objects on the table, and (6) the number and positions of lights in the scene.

The affordance detection network had 4 convolutional and corresponding transposed layers. We used KL-divergence penalty $\beta=4$ and a 10 dimensional latent space throughout the experiments. 

\subsection{Training the trajectory model}

Training data included evenly distributed samples of trajectories that moved the end-effector on top of the workspace region. The initial pose of the robot was the same as in the test environment.
Samples were generated with  $RRT^{*}$  using MoveIt!.
The value of the $\beta$ coefficient was increased while training such that it was initialized with $\beta = 10^{-8}$, and then $\beta$ was iteratively increased in 400 epoch intervals towards $\beta = 10^{-5}$. By iteratively increasing $\beta$, the model learns first to reconstruct trajectories accurately without considering the smoothness and neighborhood preservation properties of the low-dimensional space. Without this, the model got stuck to local minima, where its trajectory reconstruction accuracy was poor. 

Trajectories were encoded to 5 dimensional action space, and the number of steps in each trajectory was 24.
\vspace{-3pt}

\subsection{Policy training}
The policy training setup was constructed in simulation to mimic the physical setup (Figure \ref{fig:intro_mujoco}).
The relative poses of the camera and a cup to the robot pose corresponded to the experimental setup.
In order to train the policy, the affordance perception, policy and trajectory generation parts were combined into a single network, with the weights of the affordance perception and trajectory generation frozen. 

The target of the policy was to determine an action that generates a trajectory whose final point lies above the cup.
The training loss was calculated as the mean squared error between the target end-effector position and the actual one.

Since the trajectory is expressed in joint space, the forward kinematics solution for the final point $u_T$ was computed.
In order to backpropagate the error through this computation, the base-to-hand transformation matrix---which depends on the joint configuration ---was expressed as a PyTorch computation graph.

As the latent representation is different for various cup shapes, positions, and camera angles, we used 15 different cup shapes and randomized the camera and cup positions.
In total, our training set contained 3 million images with labeled cup positions.
The average distance error was $0.9cm$ with validation data.

\vspace{-3pt}

\subsection{End-to-end experiments}

\begin{table}[]
    \vspace{3pt}
    \caption{Experiment results for different positions in clutterless environment. All distances are given in centimeters.}
    \label{tab:resnoclut}
    \centering
    \begin{tabular}{c|c|c|c|c|c|c}
        Cup & Radius & Height & $x$ error & $y$ error & Error & Success \\
        \hline
        blue & 3.45 & 16.60 & 0.73 & 2.48 & 2.68 & 82\% \\
        can & 3.32 & 10.15 & 0.76 & 1.23 & 1.57 & 100\% \\
        can2 & 4.22 & 13.76 & 0.52 & 1.48 & 1.63 & 100\% \\
        green & 3.92 & 9.33 & 0.74 & 1.00 & 1.35 & 100\% \\
        white & 4.06 & 10.44 & 0.67 & 2.41 & 2.53 & 91\% \\
        jar & 4.75 & 16.30 & 1.52 & 3.20 & 3.84 & 82\% \\
        red & 4.30 & 8.08 & 0.59 & 1.60 & 1.76 & 100\% \\
        rocket & 4.06 & 13.06 & 0.54 & 0.99 & 1.17 & 100\% \\
        stack1 & 3.70 & 10.52 & 0.52 & 1.99 & 2.07 & 100\% \\
        stack2 & 4.46 & 8.361 & 0.56 & 3.07 & 3.14 & 75\% \\
        stack3 & 2.72 & 8.76 & 0.51 & 1.15 & 1.31 & 100\% \\
        pastel & 3.83 & 10.67 & 0.56 & 1.68 & 1.84 & 100\% \\
        yellow & 4.06 & 5.69 & 0.95 & 3.45 & 3.69 & 73\% \\
        \hline
        Average & 3.91 & 10.90 & 0.71 & 1.97 & 2.19 & 92.5\%
    \end{tabular}
\end{table}

\begin{figure}
    \centering
    \includegraphics[width=0.95\linewidth]{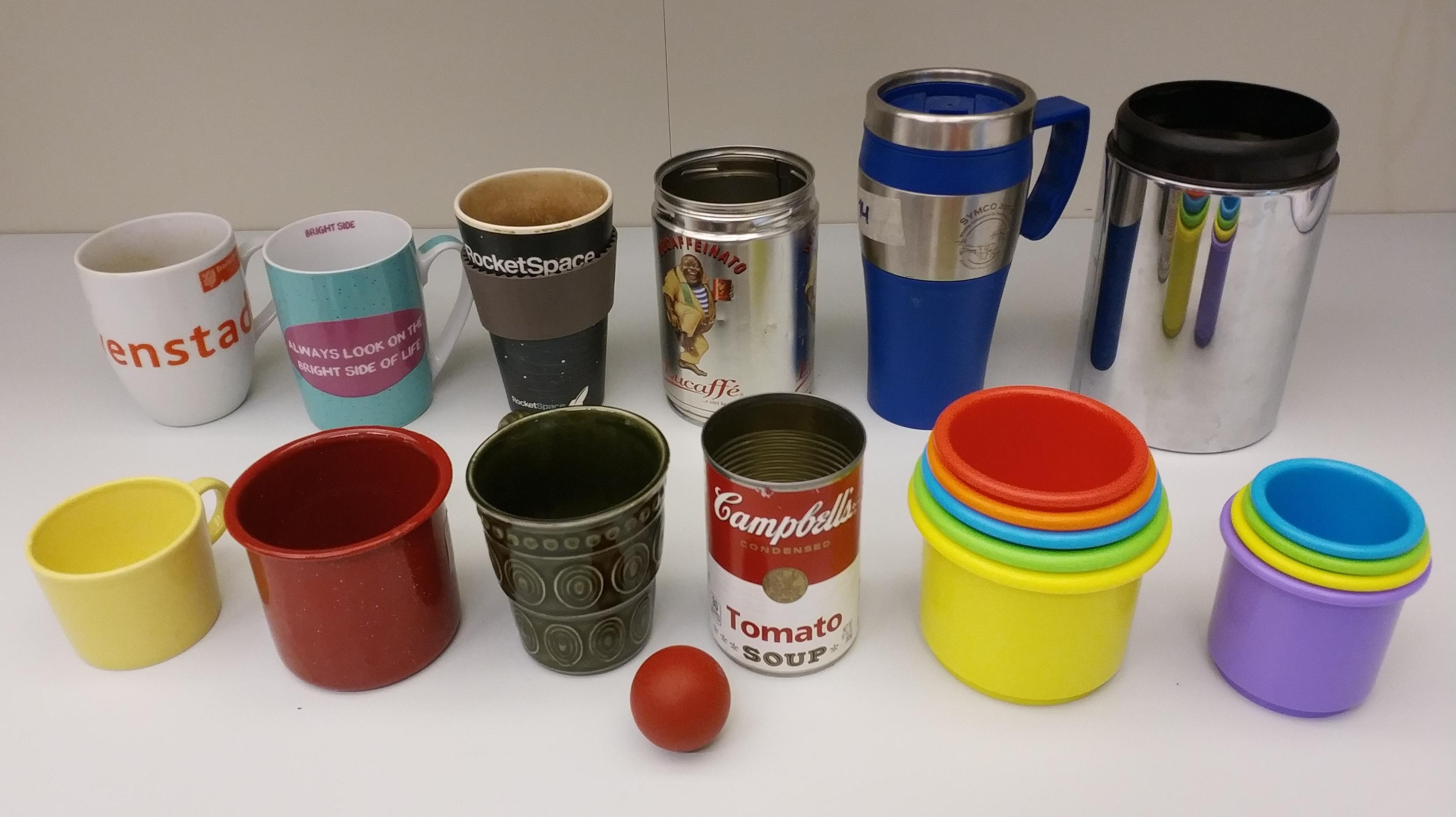}
    \caption{Various cups and the ball used throughout the experiments. Left to right, back row: white, pastel, rocket, can2, blue, jar. Front row: yellow, red, green, can, stack1, stack3. stack2 is composed from two bottom layers of stack 1, i.e. the large yellow and green parts.}
    \label{fig:my_label}
    \vspace{-14pt}
\end{figure}

We evaluated the system with and without clutter in terms of detection accuracy (position error) and task success rate (placing a ball inside the object).

\subsubsection{No clutter} Without clutter, 13 different cup like objects shown in Figure \ref{fig:my_label} were used.
Each of the objects was placed in 10 to 12 positions, with a total number of 143 trials.
The diameter of the ball was 4 cm.

The position errors and success rates are reported in Table \ref{tab:resnoclut}.
The average error over all objects was 2.2 cm with three objects having over 3 cm average error. The success rate was 100\% for 8/13 objects and 73--91\% for the rest. The accuracy can be considered good, taking into account that the system has been trained to detect any general cylindrical containers. 

The variation between objects seems to be explained by objects' appearance: The objects with the highest average distance error---yellow and jar---differ significantly from those in the training set. 
The yellow cup, for example, is smaller than ones in training set, with a height of 5.7 cm.
The system therefore expects the cup to be taller, and hence there is systematic error along the $y$ axis.
The object labeled as jar is a large aluminum container with a mirror texture on the outside.
Even though the training data did not contain any reflective objects, the system still managed to detect the opening and successfully place the ball inside of it 82\% of the times.

\begin{figure}
    \vspace{2pt}
    \centering
    \includegraphics[width=0.9\linewidth]{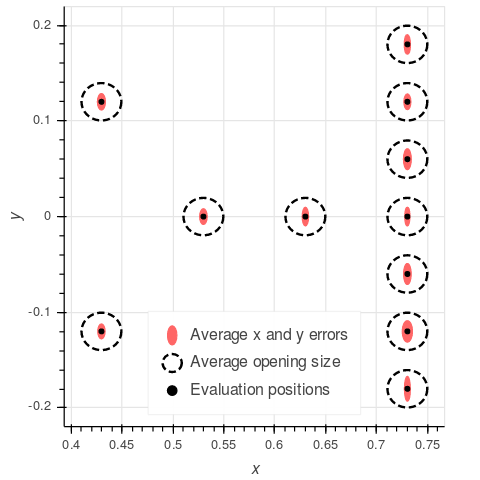}
    \caption{Evaluation positions with their respective average error ellipses marked.}
    \label{fig:pos_errors}
    \vspace{-4pt}
\end{figure}

To study the position dependency of error, all cup positions used for the experiments, together with their average error ellipses are shown in Figure \ref{fig:pos_errors}. There are no consistent differences in accuracy across positions.
However, we can see that the system performs more accurately along the $x$-axis than the $y$-axis.
The higher error along $y$ can be explained by the lack of depth information in the system and camera placement.
While the position of the object along the $x$-axis can easily be deduced from the input image, calculating the $y$-position without knowledge of depth or object size is difficult. 

\begin{table}[]
    \caption{Results of experiments in the cluttered environment. Columns $d$, \% and $n$ indicate, respectively, the average position error in cm, percentage of successful trials and the total number of trials.}
    \label{tab:my_label}
    \centering
    \begin{tabular}{c||c|c|c||c|c|c||c|c|c}
        \multirow{2}{*}{\shortstack[l]{Clutter\\ objects}} & \multicolumn{3}{c||}{Rocket} & \multicolumn{3}{c||}{Can} & \multicolumn{3}{c}{Red} \\
        \cline{2-10}
        & $d$ & \% & $n$ & $d$ & \% & $n$ & $d$ & \% & $n$ \\
        \hline
        0-6 & 1.25 & 97 & 45 & 1.64 & 92 & 28 & 2.02 & 94 & 69 \\
        7-12 & 1.86 & 96 & 30 & 2.93 & 76 & 26 & 2.4 & 82 & 64 \\
        13-18 & 2.84 & 100 & 6 & 4.47 & 73 & 30 & 4.76 & 63 & 36 \\
        19+ & 2.27 & 80 & 15 & 6.04 & 35 & 14 & 4.84 & 60 & 10
    \end{tabular}
\vspace{-3pt}
\end{table}

\begin{figure}
\vspace{-5pt}
    \centering
    \begin{subfigure}{0.48\linewidth}
        \centering
        \includegraphics[width=\linewidth]{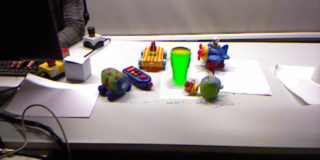}
        \caption{}
        \label{fig:litclut}
    \end{subfigure}
    \begin{subfigure}{0.48\linewidth}
        \centering
        \includegraphics[width=\linewidth]{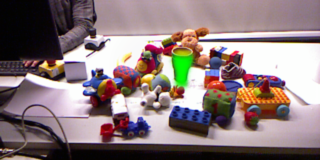}
        \caption{}
        \label{fig:lotclut}
    \end{subfigure}
    \vspace{-2pt}
    \caption{Successful affordance detection with different amounts of clutter.}
    \label{fig:cluttereverywhere}
\end{figure}

\begin{figure}
\vspace{5pt}
    \centering
    \begin{subfigure}{0.48\linewidth}
        \centering
        \includegraphics[width=\linewidth]{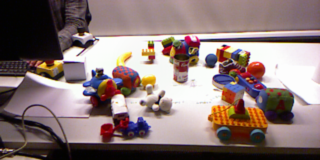}
        \caption{}
        \label{fig:distbeh}
    \end{subfigure}
    \begin{subfigure}{0.48\linewidth}
        \centering
        \includegraphics[width=\linewidth]{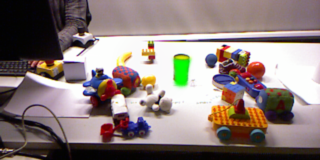}
        \caption{}
        \label{fig:distbehgone}
    \end{subfigure}
    \begin{subfigure}{0.48\linewidth}
        \centering
        \includegraphics[width=\linewidth]{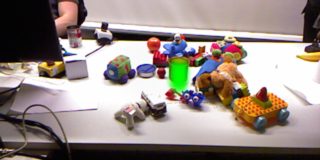}
        \caption{}
        \label{fig:distfro1}
    \end{subfigure}
    \begin{subfigure}{0.48\linewidth}
        \centering
        \includegraphics[width=\linewidth]{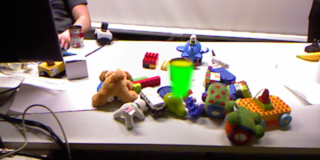}
        \caption{}
        \label{fig:distfro2}
    \end{subfigure}
    \caption{Examples of failure cases in different clutter configurations: the detection fails with clutter behind the object (\subref{fig:distbeh}) and succeeds when the clutter is removed (\subref{fig:distbehgone}). With heavy clutter in front of the object its height is incorrectly estimated (\subref{fig:distfro1}, \subref{fig:distfro2})}
    \label{fig:distractorsmessingthingsup}
\vspace{-16pt}
\end{figure}

\subsubsection{Cluttered scene} To evaluate the system's susceptibility to clutter, we chose three objects (rocket, red, can) from the ones used for previous experiments, providing a variety of different shapes, sizes and texture.
The clutter objects were toys from a toy dataset\footnote{http://irobotics.aalto.fi/software-and-data/toy-dataset}, which were not seen during training. 
They were randomly shuffled on the table between each trial and their number was steadily increased from 0 to 25.
The cup object position was fixed during these experiments.

Example scene images captured during these experiments are shown in Figure \ref{fig:cluttereverywhere}.
The \textit{wrap-grasp} affordance of the cup is visualized in green and the \textit{contain} affordance in yellow.
We can see that despite quite heavy clutter the cup was successfully identified in both situations.

The position errors and success rates are presented in Table \ref{tab:my_label}.
Not surprisingly, the errors increase and success rates decrease with increasing clutter, but the performance deterioration is gradual. Despite the training data having only 0 to 10 clutter object on the table, our system proved to be resilient to a larger amount of clutter.
This is most likely caused by applying heavy texture randomization when the training data was generated.
The perception model thus learned to perceive clutter as just a different texture of the table - both of these look identical if there is no depth information available to the model.

Noticing that the task can succeed even in  heavily cluttered scenes, such as the one shown in Figure \ref{fig:lotclut}, 
we further analyzed the data to determine failure cases. 
We found that the amount of clutter was not as important as its positioning---when clutter objects were located directly in front of the cup, covering its bottom part, the system quite often failed to correctly position the arm on the $y$ axis.
This is expected to happen---when the bottom part of the cup is not visible, it is impossible for the model to deduce its height, and thus estimate the $y$ position.
The situation is illustrated in Figures \ref{fig:distfro1} and \ref{fig:distfro2}. 

Putting object directly behind the cup opening also had a noticeable impact on performance, although it was less significant.
In that case, the problems are likely to be caused by the perception model having problems estimating where the opening of the cup ends.
It is, however, a difficult task to perform without depth data or stereovision.
This is illustrated in Figure \ref{fig:distbeh}, where a distractor object is placed directly behind the opening of a can.
As the distractor and the inside of the can have a similar color, it becomes difficult to identify the opening in the image, and so the affordance part fails.
When the distractors are removed from the back (Figure \ref{fig:distbehgone}), the object is successfully identified again.

\vspace{-1pt}
\section{Conclusions and future work}
\vspace{-1pt}

We studied the use of latent space affordance encodings in training manipulator policies. We demonstrated that the affordance encodings can be used to train policies successfully.
We also showed that those latent space representations, if trained in a randomized domain, can be made invariant to, e.g., distractor objects and textures, making them applicable in zero-shot sim-to-real transfer.

We also developed a method for generating a domain-randomized affordance dataset, which works across various domains (heavily randomized images from Blender, simplistic frames from MuJoCo and real world images with different amounts of clutter).
The method is easily scalable to new types of objects and affordances, and the software to generate the dataset has been released to the public.

The proposed methods operate on bare RGB images without depth information. 
Many recent affordance detection methods use depth as an additional cue. 
However, domain randomization is an essential component for the simulation-to-real transfer. 
Thus, domain randomization for depth data seems to be an appealing open research problem. 

Another recent idea applied in affordance detection is to include semantic information from e.g.~object recognition. 
These methods do not typically have encoder-decoder structure so that the modular approach proposed here could not be directly applied to find corresponding latent representations. 
Finding such representations for more complex affordance detectors would be useful and remains an open problem. 


Finally, the proposed approach is based on feedforward trajectory generation. 
However, the latent affordance representations could be used also as the state space of feedback policies, for example, to learn closed-loop visual alignment tasks, or even active closed-loop search strategies for objects with particular affordances.

\vspace{-4pt}

\addtolength{\textheight}{-12cm}   







\bibliographystyle{unsrt}
\bibliography{main}

\end{document}